\documentclass[11pt]{article}
\usepackage[final]{acl}

\usepackage{times}
\usepackage{latexsym}
\usepackage{amsmath} 
\usepackage[T1]{fontenc}

\usepackage[utf8]{inputenc}

\usepackage{microtype}

\usepackage{inconsolata}
\usepackage{subcaption}

\usepackage{booktabs}
\usepackage{multirow}
\usepackage{graphicx}
\usepackage{makecell}
\usepackage{tabularx}
\usepackage[most]{tcolorbox}
\usepackage{amsmath, amssymb}
\usepackage{xcolor}
\usepackage{cuted}

\usepackage{array, xcolor, colortbl}
\usepackage{float} 

\usepackage{ragged2e}
\usepackage{caption}
\usepackage{amssymb}
\usepackage[T1]{fontenc}
\definecolor{codeblue}{RGB}{0, 82, 147}     
\definecolor{codegreen}{RGB}{0, 128, 0}    
\definecolor{codegray}{RGB}{100, 100, 100}  
\definecolor{codeorange}{RGB}{230, 145, 56} 
\definecolor{darkerblue}{rgb}{0,0.08,0.45}
\definecolor{royalblue}{RGB}{65,105,225}
\definecolor{lightblue}{RGB}{221,235,247}
\definecolor{fig3blue}{RGB}{47, 122, 232}  
\definecolor{fig3red}{RGB}{213, 32, 52}
\definecolor{fig3green}{RGB}{0, 137, 72} 
\definecolor{fig3yellow}{RGB}{217, 161, 5}
\definecolor{gray94}{gray}{.94}
\definecolor{gray90}{gray}{.90}

\definecolor{darkgreen}{RGB}{34,139,34}

\newcolumntype{g}{>{\columncolor{gray94}}c}
\newcolumntype{b}{>{\columncolor{lightblue}}c}

\newcommand{\bcell}[1]{\cellcolor{gray94}{#1}}

\usepackage{caption}
\title{Reasoning Over Space: Enabling Geographic Reasoning for LLM-Based Generative Next POI Recommendation}

\author{
 \textbf{Dongyi Lv\textsuperscript{1,2}\thanks{Work done during the internship at Amap, Alibaba Group.}},
 \textbf{Qiuyu Ding\textsuperscript{2}},
 \textbf{Heng-Da Xu\textsuperscript{2}},
 \textbf{Zhaoxu Sun\textsuperscript{2}},
\\
 \textbf{Zhi Wang\textsuperscript{1}\thanks{Corresponding author.}},
 \textbf{Feng Xiong\textsuperscript{2}},
 \textbf{Mu Xu\textsuperscript{2}},\\
 \textsuperscript{1}School Of Software Engineering, Xi’an Jiaotong University,
 \textsuperscript{2}Amap, Alibaba Group
\\\texttt{lvdongyi@stu.xjtu.edu.cn, zhiwang@xjtu.edu.cn,}\\
   \texttt{\{qiuyu.dqy,xuhengda.xhd,szx430629,xf250971,xumu.xm\}@alibaba-inc.com}}

\begin{document}
\newcommand\symbolfootnote[2]{%
  \begingroup
  \renewcommand\thefootnote{\ensuremath{#1}}\footnote{#2}%
  \addtocounter{footnote}{-1}%
  \endgroup
}
\maketitle
\begin{abstract}
Generative recommendation with large language models (LLMs) reframes prediction as sequence generation, yet existing LLM-based recommenders remain limited in leveraging geographic signals that are crucial in mobility and local-services scenarios. Here, we present \textsc{Reasoning Over Space} (ROS), a framework that utilizes geography as a vital decision variable within the reasoning process. ROS introduces a Hierarchical Spatial Semantic ID (SID) that discretizes coarse-to-fine locality and POI semantics into compositional tokens, and endows LLM with a three-stage Mobility Chain-of-Thought (CoT) paradigm that models user personality, constructs an intent-aligned candidate space, and performs locality informed pruning. We further align the model with real world geography via spatial-guided Reinforcement Learning (RL). Experiments on three widely used location-based social network (LBSN) datasets show that ROS achieves over 10\% relative gains in hit rate over strongest LLM-based baselines and improves cross-city transfer, despite using a smaller backbone model.\symbolfootnote{\ddagger}{Code available at \url{https://github.com/amap-cvlab/Reasoning-Over-Space}}
\end{abstract}

\section{Introduction}

Generative recommendation with large language models (LLMs)  \cite{rajput2023recommender,zhai2024hstu,senel2024generative,qu2025tokenrec,gao2024generative,lee2025gram} demonstrate a strong, promising alternative to conventional retrieval and ranking pipelines in practice. By framing recommendation as sequence generation, these models can be trained end-to-end, and often benefit from scaling in both data and model capacity\cite{deng2025onerec,chen2024hllm,han2025mtgr}, which improves overall generalization and reduces reliance on task specific architectures.

\begin{figure}
    \centering
    \includegraphics[width=\linewidth]{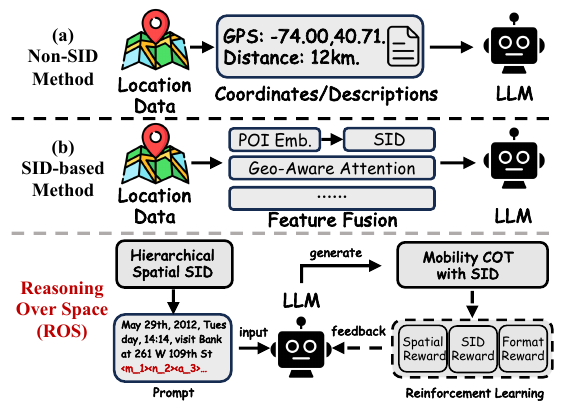}
    \caption{Comparison of different paradigms for incorporating geographic signals into LLM recommendation.}
    \label{fig:differentgeofusion}
\end{figure}

Without explicitly modeling geographic signals, a recommender struggles to capture core mobility patterns, such as users preference for short distance transitions and the frequent consecutiveness of spatially proximate Point-of-Interests\footnote{A POI is a place that a user can visit, typically represented by its geographic coordinates and semantic metadata.}(POIs). It also fails to exploit spatiotemporal feasibility priors, most notably that long distance moves are unlikely within short time intervals, which can lead to geographically implausible recommendations.

Despite their promise, current LLM-based generative recommenders remain limited in incorporating geographic signals that matter in mobility and local-services\cite{zhang2025survey,feng2024move,zhong2025comapoi}. 
In mobility and local-services scenarios, next POI recommendation seeks to generate the next visited POI given a user’s chronologically ordered check in history.
Existing work typically incorporates location either by converting it into text inputs, such as appending coordinates and POI descriptions \cite{feng2024move}, treating it as auxiliary numeric features \cite{wang2025generative,liu2025geography,jiang2025llm} that are fused into representations, or by fusing location via geo-aware self attention\cite{wei2025oneloc}, see Figure~\ref{fig:differentgeofusion}. While these designs expose location context to the model, they do not leverage the model’s reasoning capability to treat geography as a first class decision variable, making it difficult to compare candidates spatially, assess travel feasibility, and rule out geographically implausible options.

To bridge this critical gap, we propose a novel framework Reasoning Over Space (ROS) that incorporate geographical information in reasoning itself. Specifically, we (1) Propose a Hierarchical Spatial Semantic ID (SID) that encodes coarse-to-fine locality together with functional semantics as compositional tokens;(2) A recommender model is trained to apply a structured three stage reasoning paradigm that explicitly uses locality constraints for candidate elimination; and (3) Align the model with real world geography via spatial-guided Reinforcement Learning (RL).


\section{Methodology}
\begin{figure*}[t]
    \includegraphics[width=1.0\linewidth]{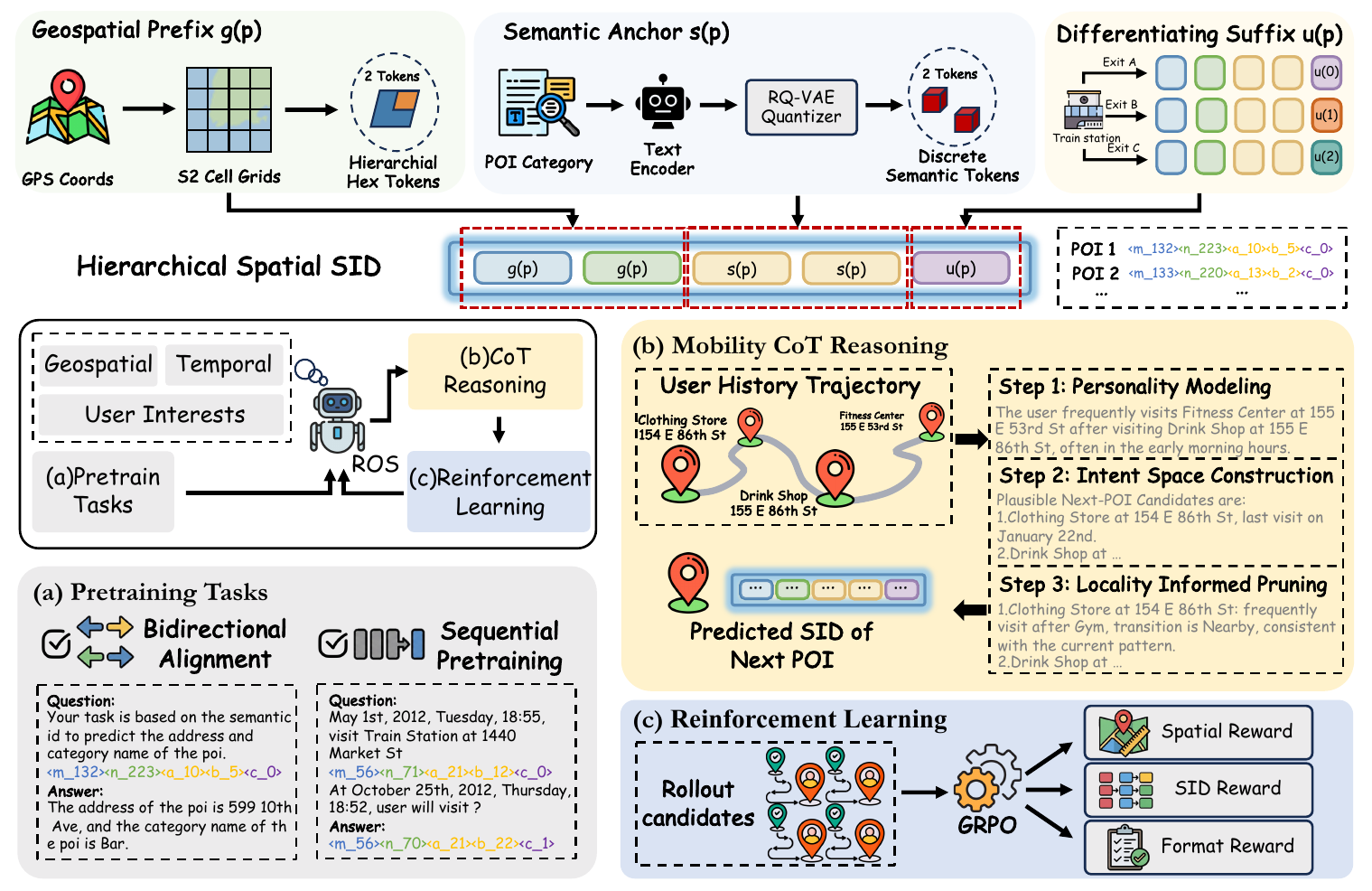}
    \caption{Overall Framework of our method.}
    \label{fig:overall}
\end{figure*}

In this section, we first formulate the next POI recommendation problem, then introduce a Hierarchical Spatial SID for structured POI representation, describe a three-stage Mobility CoT reasoning paradigm, and finally detail our spatial-guided RL procedure. The overall framework of ROS is presented in Figure \ref{fig:overall}.
\subsection{Problem Formulation.}
Next POI recommendation aims to predict a user’s next visited POI from their historical check-in behavior in location-based social networks (LBSN). For each user, we observe a chronologically ordered check-in trajectory
$\mathcal{H}=\{(p_i,t_i,c_i,\mathbf{x}_i)\}_{i=1}^{n}$,
where $p_i$ denotes the visited POI, $t_i$ is the timestamp, $c_i$ is the POI category,
and $\mathbf{x}_i\in\mathbb{R}^2$ is the GPS coordinate of $p_i$.
Given history trajectories of users $\mathcal{H}$, the goal is to predict the next POI $p_{n+1}$ that the user will visit.
Each query is associated with a unique ground truth next POI, and the recommendation is
considered correct if $p_{n+1}$ exactly matches the ground truth.

\subsection{Spatial Semantic Tokenization}
\subsubsection{Hierarchical Spatial SID}

Human mobility\footnote{Human mobility refers to the time resolved movement of individuals across geographic locations, commonly represented as a spatiotemporal trajectory of visited places and displacements\cite{gonzalez2008understanding}.} is governed not merely by temporal correlations, but by a latent structure that couples geography, semantics, and behavioral regularities. Some recent works introduce SID of POIs, but they still fail to exploit fine grained spatial structure while remaining high collision rate, weakening the model’s ability to reason about mobility.
To reveal and exploit the latent spatial semantic structure underlying human mobility, we construct a hierarchical spatial SID that transforms the raw POI into a symbolic, interpretable spatial semantic coordinate system, enabling LLMs to reason over location, function, and geographic granularity in a discrete and compositional manner. We decompose each POI $p$ into three components:
\[
\text{SID}(p) = [g(p)\ ;\ s(p)\ ;\ u(p)]
\]
where $g(p)$ encodes geospatial locality, $s(p)$ captures functional semantics, 
and $u(p)$ is a lightweight discriminator that ensures uniqueness within each 
spatial semantic class.

\noindent\textbf{Geospatial Prefix.}
Each POI coordinate is converted into an S2 Cell Id\cite{s2geometry} and represented as a fixed 
hex string. Then we remove a dataset wide shared prefix and keep the next $2\times B$ hex 
digits, which are grouped into $B$ tokens:
\[
g(p)=\big[\tilde{h}_{1:2},\ldots,
         \tilde{h}_{2B-1:2B}\big]
\]
Higher order tokens correspond to larger spatial regions, inducing a natural 
hierarchy in the discrete prefix that nearby POIs tend to share similar prefix.
This hierarchy later allows the model to receive partial reward for predicting the correct coarse region even when the exact POI is wrong, and it naturally supports region level generalization.

\noindent\textbf{Semantic Anchor.}
To encode functional roles, we embed each POI's category description using \texttt{Qwen-0.6B} and quantize the embedding via RQ-VAE\cite{lee2022autoregressive}. The resulting discrete tokens $s(p)$ clusters POIs with similar semantics, making semantic similarity explicitly accessible to the language model.

\noindent\textbf{Differentiating Suffix.}
The suffix $u(p)$ carries no additional semantics, it only disambiguates POIs 
that share the same prefix and semantic anchor.

Together, the three components turn opaque POI IDs into an interpretable, hierarchical, and uniquely identifying code. This unified representation makes it easier for our model to acquire spatial semantic regularities during pretraining, and later provides a structured space on which we define SID level consistence rewards in reinforcement learning.

\subsubsection{SID Grounding Pretraining}

After constructing the SID representation, we pretrain the model to understand and utilize the SID hierarchy leveraging two objectives.

\noindent\textbf{Bidirectional Text SID Alignment.}
We align each SID with its category and address via supervised fine-tuning (SFT) in both directions: the model is trained to predict the SID from text and to generate text from the SID. This bidirectional alignment grounds SID tokens in functional semantics and fine-grained neighborhood cues.

\noindent\textbf{Sequence Pretraining for Mobility.}
In order to expose the model to temporal, semantic, and coarse spatial regularities such as recency, category repetition, and neighborhood continuity, we further train the model on an auto-regressive next POI recommendation task.

While this pretraining stage exposes the model to spatial semantic structure and generic mobility patterns, it still treats next POI recommendation as a single step mapping. In the next section, we explicitly structure the recommendation as a multi step reasoning process.

\subsection{Mobility Chain-of-Thought Reasoning}

Human mobility is not a flat classification problem over POI IDs, but a sequence of constrained decisions driven by time, intent, and geography. To reflect this, we model next POI recommendation as an explicit reasoning process rather than a one-shot label prediction.
Instead of mapping a trajectory to the next POI in a single opaque step, we fine-tune the LLM to generate Mobility CoT traces that decompose the decision into three deterministic stages. 
Starting from the most recent check-ins, the model progressively explores, evaluates, and narrows down plausible candidates.

To make geographic reasoning accessible in natural language, we serialize each POI with its category name and street-level address. Compared with arbitrary POI IDs, street addresses provide fine-grained neighborhood cues, allowing the model to naturally perceive spatial locality and recurring areas directly from the given information.

\noindent\textbf{Personality Modeling.}
In order to ground the recommendation in the user’s mobility persona, we begin by distilling the most recent check-ins into a compact, evidence-based profile.
Scanning the trajectory backward, the model extracts factual statements about last visit times, weekend or holiday indicators, repeated category or regional patterns, and recent movement characteristics. Since POIs are represented with both its category name and street-level address, the extracted evidence can also highlight locality patterns, which further supports geographic plausibility judgments in later steps.

\noindent\textbf{Intent Space Construction.}
Next, the model aligns the candidate space with the user’s intent by generating plausible next POI candidates conditioned on recent behavioral evidence. We encourage model to generate a recency-first candidate space, guiding the model to prioritize more recently visited POIs before older ones. This inductive bias introduces an implicit temporal structure that captures both short-term intent and habitual preferences, while also removes spurious positional cues introduced during data synthesis, reducing bias from arbitrary candidate permutations.

\noindent\textbf{Locality Informed Pruning.}
In order to make geography a vital decision criterion, we finally evaluate and prune candidates using both locality cues and feasibility signals. Street-level addresses provide neighborhood-level anchors that allow the model to detect whether a candidate stays within the user’s active area, while precomputed transition distances in the prompt offer a constraint on movement plausibility. By jointly checking temporal consistency, semantic fit, and spatial feasibility, the model deterministically discards incompatible candidates and retains only those that satisfy the extracted constraints.

\medskip
Using a teacher model to generate Mobility CoT traces as supervision, the student model learns to emulate this three-step paradigm, enabling a structured and reproducible reasoning process for next POI recommendation rather than depending on shallow information from trajectories.

\subsection{Spatial-Guided Reinforcement Learning}
\label{sec:rl}

Supervised CoT training enable our model apply the three step Mobility CoT paradigm and verbalize its reasoning, but it does not by itself guarantee geographically faithful recommendations. To further align the model with real world geography and the hierarchical SID space, we apply spatial-guided RL with a composite reward that jointly encourages geographically plausible recommendations, SID level correctness, and format consistent reasoning.

\subsubsection{Spatial Grounding Reward}

Let $d$ be the haversine distance\cite{sinnott1984virtues} between predicted and ground-truth POIs. We transform it as $\tilde d = \log(1 + d)$ and map $d$ to a clipped linear reward:
\[
r_{\text{dist}}(d)
= \operatorname{clip}\!\Bigl(
r_{\max}+\kappa(\tilde d-\tilde d_{\text{near}}),\, r_{\min},\, r_{\max}
\Bigr)
\]
where $\kappa$ is the slope between two thresholds $\tilde d_{\text{near}}$ and $\tilde d_{\text{far}}$\footnote{We set $d_{\text{near}}{=}0.1$\,km and $d_{\text{far}}{=}3.0$\,km.
These thresholds align with common practice in trajectory and POI mining for defining local tolerance, and trip and itinerary recommendation for neighborhood scale pruning~\cite{zheng2015trajectory,yeow2021point,gao2022self,halder2022efficient}.}. The \textit{log transform} makes the reward sensitive to local deviations while saturating penalties for very distant recommendations.

\subsubsection{Hierarchically Weighted SID Reward}

Each POI is represented by SID, which composed by geospatial prefix $g(p)$, semantic anchor $s(p)$, and differentiating suffix $u(p)$.
\[
    \text{SID}(p) = [g(p); s(p); u(p)]
\]
We decompose the geospatial prefix and semantic anchor into hierarchical sub-tokens and assign monotonically decreasing weights along the hierarchy, so that coarse location and category matches contribute more than fine-grained matches. The resulting score defines a base correctness reward $r_{\text{base}}$. Crucially, we further introduce a hard exactness bonus $\lambda_u$ when the entire SID, including the suffix, is predicted exactly:
\[
  r_{\text{acc}} = \min\{1,\, r_{\text{base}} + \lambda_u \mathbb{I}[\mathrm{SID} = \widehat{\mathrm{SID}}]\}.
\]
The format reward $r_{\text{fmt}}$ is a binary signal that simply checks whether the generated trace conforms to the prescribed three step CoT paradigm.

    \subsubsection{Unified Reward Optimization}
To jointly encode geographic preference, symbolic correctness, and format adherence into a single training signal, for each generated trajectory trace, we define the overall reward as:
\[
r = r_{\text{fmt}} + \alpha  r_{\text{acc}} + \beta r_{\text{dist}}.
\]
Therefore, geographic preference, SID level symbolic correctness, and structured reasoning are optimized in a unified way. We then optimize the model with Group Relative Policy Optimization (GRPO)\cite{shao2024deepseekmath}.
As a result, the optimized model internalizes geography and SID hierarchy as actionable preferences, yielding recommendations that are not only symbolically correct but also spatially plausible, while maintaining trace-to-decision consistency under the three-stage Mobility CoT paradigm.

\section{Experiments}

\newcolumntype{L}{>{\centering\arraybackslash}p{1.6cm}}

\newcolumntype{R}{>{\centering\arraybackslash}p{1.6cm}}

\begin{table}[t]
    \centering
    \caption{\textbf{Next POI recommendation performance.} All results are HR@1.
    }
    \label{tab:main-results-redesign}

    \resizebox{\linewidth}{!}{%
    \begin{tabular}{llccc}
        \toprule
        \rowcolor{gray94}
        \textbf{Category} & \textbf{Method} & \textbf{NYC} & \textbf{TKY} & \textbf{CA} \\
        \midrule

        \multirow{1}{*}{Traditional}
            & PRME  & 0.1159 & 0.1052 & 0.0521 \\
        \midrule

        \multirow{5}{*}{Neural-based}
            & GETNext & 0.2435 & 0.1829 & 0.1357 \\
            & TPG     & 0.2555 & 0.1420 & 0.1749 \\
            & MTNext  & 0.2620 & 0.2575 & 0.1453 \\
            & STHGCN  & 0.2734 & 0.2950 & 0.1730 \\
            & ROTAN   & 0.3106 & 0.2458 & 0.2199 \\
        \midrule

        \multirow{8}{*}{LLM-based}
            & ST-GR     & 0.2920 & 0.2610 & 0.1659 \\
            & LLM4POI   & 0.3372 & 0.3035 & 0.2065 \\
            & GNPR-SID  & 0.3618 & 0.3062 & 0.2403 \\
            & GA-LLM    & 0.3919 & 0.3482 & 0.2566 \\
            & CoAST     & 0.4027 & 0.3310 & 0.2721 \\
            
            \rowcolor{gray94}
            & ROS & \textbf{0.4478} & \textbf{0.3864} & \textbf{0.3149} \\

            \rowcolor{gray94}
            & \textit{vs.\ SOTA} ($\Delta$\%) &
            {\footnotesize\textcolor{green!60!black}{\textbf{+11.2\%}$\uparrow$}} &
            {\footnotesize\textcolor{green!60!black}{\textbf{+11.0\%}$\uparrow$}} &
            {\footnotesize\textcolor{green!60!black}{\textbf{+15.7\%}$\uparrow$}} \\
        \bottomrule
    \end{tabular}
    }
    \label{tab:main-results}
\end{table}

\subsection{Experimental Setting}

We thoroughly evaluate our method on three widely used LBSN benchmarks: Foursquare-NYC, Foursquare-TKY \cite{foursquarenyctky}, and Gowalla-CA\cite{gowallaca}. We follow the LLM4POI\cite{li2024large} preparation pipeline: (i) Filter users/POIs with fewer than 10 check-ins; (ii) Sort check-ins chronologically and segment trajectories using a time-gap threshold; (iii) Split data into 80\%/10\%/10\% train/validation/test sets in temporal order while keeping validation/test users and POIs observed in training. We additionally apply reverse geocoding\cite{LI201895} to obtain street-level addresses as auxiliary textual cues. For each experiment, we run inference for three times and report the mean score.

\paragraph{Baselines}
We compare our method against three categories of baselines: (i) Traditional baseline  PRME~\cite{feng2015personalized};
(ii) Neural-based baselines include GETNext~\cite{yang2022getnext}, TPG~\cite{luo2023timestamps}, MTNext~\cite{hu2018mtnet}, STHGCN~\cite{yan2023spatio}, and ROTAN~\cite{feng2024rotan}; and (iii) LLM-based baselines include SpaceTime-GR~\cite{lin2025spacetimegrspacetimeawaregenerativemodel}, LLM4POI~\cite{li2024large}, GNPR-SID~\cite{wang2025generative}, GA-LLM~\cite{liu2025geography}, and CoAST~\cite{zhai2025cognitive}.
Detailed descriptions of these baselines are provided in Appendix~\ref{app:baselines}.

\paragraph{Evaluation Metrics} HitRate@1 (HR@1) is employed as the main evaluation metric. In our settings, each request is associated with a unique ground-truth POI, and the model recommends exactly one POI. Under this one-to-one recommendation regime, HR@1 directly measures whether the model makes a correct decision, providing an unambiguous and faithful reflection of recommendation accuracy.

\paragraph{Implementation Details}
All experiments are conducted on 8$\times$NVIDIA H20 GPUs. We initialize the student model from \texttt{Qwen3-4B}~\cite{yang2025qwen3}. \texttt{Qwen3-235B-A22B} is employed as a teacher to generate three-stage CoT traces, and fine-tune the 4B student on these traces for 2 epochs. Additional training and hyperparameter details are provided in the appendix \ref{app:detail}.

\subsection{Main Results}

Table~\ref{tab:main-results} reports next POI recommendation performance on three LBSN benchmarks. Despite using a compact 4B student backbone, our method consistently outperforms the strongest LLM-based baselines that are built on 7B models (CoAST and GA-LLM). On NYC and CA, we surpass CoAST with relative HR@1 gains of $+11.2\%$ and $+15.7\%$, respectively; on TKY, we improve over GA-LLM by $+11.0\%$ HR@1. Notably, these improvements are achieved without resorting to backbone scaling. By turning mobility into a compositional spatial-semantic decision process, the model learns a stronger notion of “where” and “why” behind the next visit. This suggests that structured spatial semantics and explicit reasoning supervision can unlock more effective generalization than merely enlarging the backbone.

\subsection{Ablation Study}

\paragraph{Analysis of SID Expressivity.}
\begin{table}[t]
    \centering
    \caption{\textbf{Ablation on SID design and pretraining tasks.}
    T1: sequence modeling pretraining; 
    T2: text-SID alignment pretraining.
    All results are HR@1.}

    \newcolumntype{M}{>{\raggedright\arraybackslash}p{1.8cm}}

    \resizebox{\linewidth}{!}{%
    \begin{tabular}{M|cc|RRR}
        \toprule
        \rowcolor{gray94}
         \textbf{SID type} & \textbf{T1} & \textbf{T2} & \textbf{NYC} & \textbf{TKY} & \textbf{CA} \\
        \midrule
        Traditional & \checkmark &        & 0.3704 & 0.3310 & 0.2608 \\
        HS-SID   & \checkmark &        & 0.3877 & 0.3338 & 0.2629 \\
        HS-SID   & \checkmark & \checkmark & \bcell{\textbf{0.3918}} & 
                                              \bcell{\textbf{0.3372}} &
                                              \bcell{\textbf{0.2657}} \\
        \bottomrule
    \end{tabular}
    }

    \label{tab:ablation-sid}
\end{table}
As shown in Table \ref{tab:ablation-sid}, we first replace our hierarchical spatial SID with traditional non-hierarchical SID while keeping all prompts and training settings identical.  
Compared with traditional SID, our hierarchical spatial SID yields a clear improvement on NYC and comparable performance on TKY and CA, indicating that the additional geospatial granularity primarily benefits regions with higher spatial density.  
Incorporating the text-SID alignment objective further improves model performance across all three datasets, showing that grounding discrete SID tokens in natural language descriptions enhances functional discrimination.

\paragraph{Analysis of Mobility CoT.}
\begin{table}[t]
    \centering
    \caption{\textbf{Ablation on Mobility CoT stages.}
    Step 1: personality modeling; Step 2: intent space construction; Step 3: locality informed pruning; 
    Each row enables a subset of stages and reports HR@1.}
    \label{tab:ablation-stages-nyc}
    \resizebox{0.8\linewidth}{!}{
        \begin{tabular}{c c c | c c}
            \toprule
            \rowcolor{gray94}
            \textbf{Step 1} & \textbf{Step 2} & \textbf{Step 3} & \textbf{NYC} & \textbf{Gain($\Delta$\%)} \\
            \midrule
             &  &  & 0.3918 & -- \\
            \midrule
             
            \checkmark &  &  & 0.4105 & +4.8\% \\
             & \checkmark &  & 0.4098 & +4.6\% \\
             &  & \checkmark & 0.4029 & +2.8\% \\
            \midrule
             
             & \checkmark & \checkmark & 0.4119 & +5.1\% \\
            \checkmark &  & \checkmark & 0.4091 & +4.4\% \\
            \checkmark & \checkmark &  & 0.4140 & +5.7\% \\
            \midrule
            
            \checkmark & \checkmark & \checkmark & \cellcolor{gray94} \textbf{0.4181} & \cellcolor{gray94}\textbf{+6.7\%} \\
            \bottomrule
        \end{tabular}
    }
\end{table}

Table~\ref{tab:ablation-stages-nyc} indicates the necessity of each Mobility CoT stage. While any single stage improves over direct recommendation, the best result is achieved only when all three stages are composed. Suggesting that Mobility CoT is not a collection of independent heuristics but a coupled reasoning scaffold, \textsc{Personality Modeling} distills trajectory evidence into stable constraints, \textsc{Intent Space Construction} organizes a structured candidate space under these constraints, and \textsc{Locality Informed Pruning} provides the decisive verification signal that eliminates candidates inconsistent with the extracted evidence. Removing any stage weakens one link of this consecutive CoT, leading to a drop even when the remaining stages are present.

\begin{table}[t]
    \centering
    \caption{\textbf{Perturbing geographic cues after RL.}
    We report HR@1 on three datasets.}
    \label{tab:geo-cue-perturb}
    \resizebox{0.8\linewidth}{!}{%
    \begin{tabular}{l | c c c}
        \toprule
        \rowcolor{gray94}
        \textbf{Variant} & \textbf{NYC} & \textbf{TKY} & \textbf{CA} \\
        \midrule
        Full model &\cellcolor{gray94} \textbf{0.4478} & \cellcolor{gray94}\textbf{0.3864} & \cellcolor{gray94}\textbf{0.3149} \\
        Random Address & 0.4347 & 0.3721 & 0.2971 \\
        W/O Address & 0.4195 & 0.3444 & 0.2842 \\
        W/O Distance & 0.4416 & 0.3849 & 0.3139 \\
        \bottomrule
    \end{tabular}}
\end{table}
To better investigate how geographic cues support the above paradigm, we perturb the POI representation and distance annotations at inference time after reinforcement learning, as summarized in Table~\ref{tab:geo-cue-perturb}. Replacing each POI address with a random address leads to a consistent decline, indicating that street-level addresses act as a native locality prior that helps the model relate candidates through shared areas. Removing addresses entirely causes a larger degradation, consistent with the fact that our CoT relies on category name and street-level address as an anchor to disambiguate POIs during candidate pruning. Finally, removing explicit transition distances also hurts performance,  suggesting that distances provide a complementary feasibility prior that sharpens the verifier when ruling out implausible long jumps, beyond what can be inferred from address tokens alone.

\paragraph{Analysis of Spatial-Guided RL.}
\begin{table}[t]
    \centering
    \caption{\textbf{Effect of correctness bonus weight $\lambda_u$.}
    All results are HR@1.}
    \label{tab:bonus-lambda}
    \resizebox{0.8\linewidth}{!}{
        \begin{tabular}{c|ccccc}
            \toprule
            \rowcolor{gray94}
            \textbf{$\lambda_u$} & 0.0 & 0.1 & 0.3 & 0.5 & 1.0 \\
            \midrule
            NYC & 0.4423 & \textbf{0.4478} & 0.4444 & 0.4395 & 0.4381 \\
            \bottomrule
        \end{tabular}
    }
\end{table}

\begin{figure}[t]
    \centering
    \includegraphics[width=1.0\linewidth]{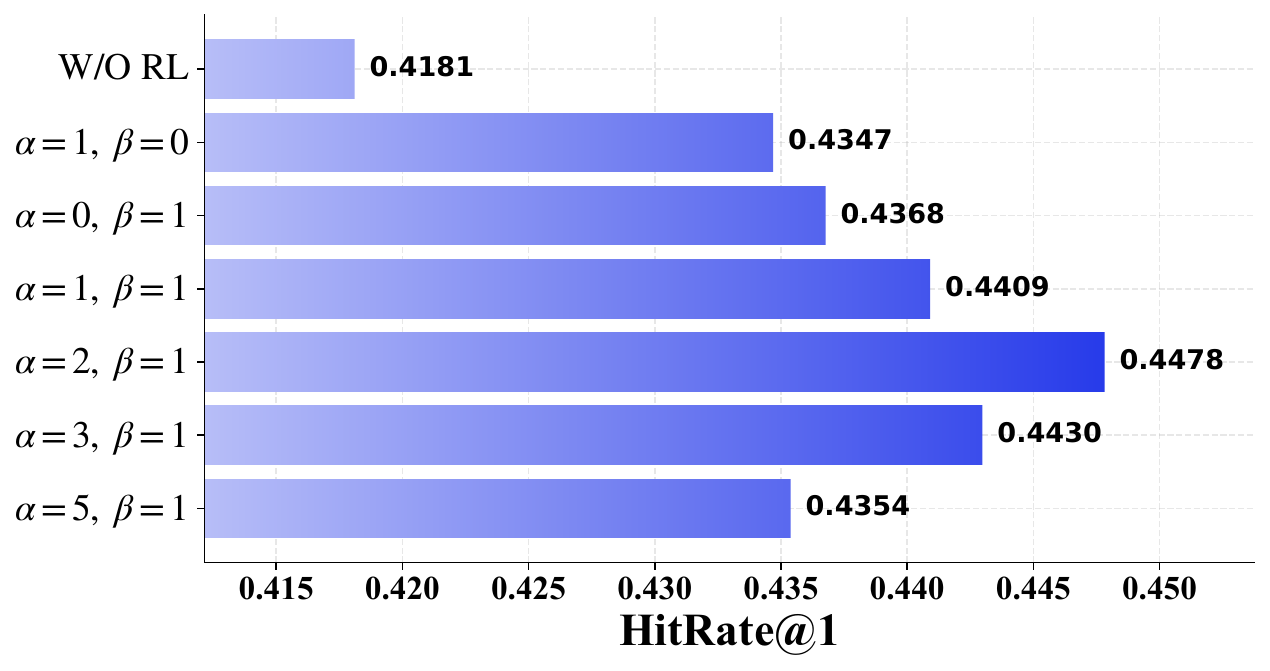}
   \caption{\textbf{Impact of reward weighting in spatial-guided RL.}
$\alpha$ and $\beta$ indicate the weights of the hierarchical correctness reward and the distance-based grounding reward, respectively. Results are reported in HR@1 on NYC.}
    \label{fig:reward-weight}
\end{figure}

Spatial-guided RL consistently improves next POI generation beyond supervised Mobility CoT fine-tuning, indicating that RL provides an additional alignment stage that better reflects mobility constraints. Figure~\ref{fig:reward-weight} further shows that using either hierarchical correctness or distance grounding alone already brings noticeable gains, while combining them yields the strongest performance. This suggests the two signals are complementary, $r_{\text{acc}}$ encourages SID-level semantic correctness, whereas $r_{\text{dist}}$ injects a spatial-guided preference that discourages geographically implausible candidates. Importantly, the weighting study reveals a clear trade-off, moderate emphasis on correctness is beneficial, but overly large $\alpha$ degrades performance, likely because discrete SID matching starts to dominate optimization and weakens the influence of spatial constraints during candidate discrimination. 

Table~\ref{tab:bonus-lambda} further analyzes the correctness bonus weight $\lambda_u$, where a small bonus ($\lambda_u{=}0.1$) yields the best HR@1, while larger values gradually hurt performance. This indicates that a small exact match bonus can help training by sharpening the preference for fully correct predictions, whereas an excessive $\lambda_u$ makes the optimization overly reliant on the sparse all correct signal. Based on these sensitivity analyses, we adopt $\alpha{=}2$, $\beta{=}1$, and $\lambda_u{=}0.1$ as robust defaults that balance semantic correctness with geographic plausibility.

\subsection{Generalization Analysis}
\paragraph{Cross City Generalization.}
\begin{figure*}[t]
  \centering
  \includegraphics[width=\linewidth]{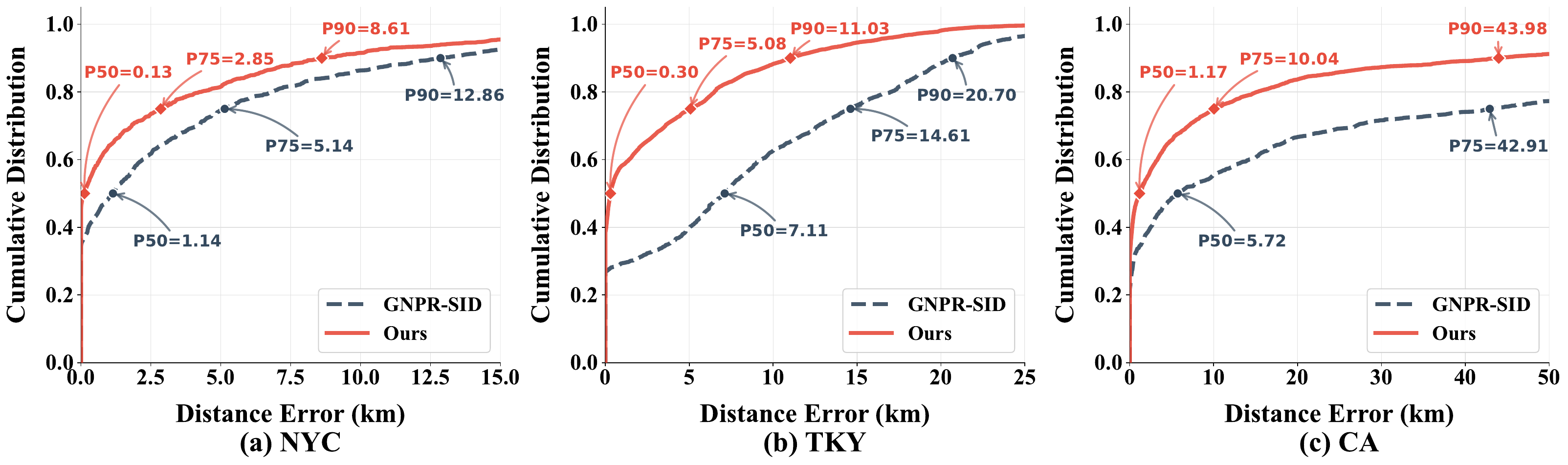}
  \caption{\textbf{Cumulative Distribution of three datasets.} Larger area under curve indicating better performance.}
  \label{fig:dist-cdf}
\end{figure*}
\begin{table}[t]
    \centering
    \caption{\textbf{Cross-city generalization performance.}
    Models are trained on one city and directly evaluated on other cities.
    All results are HR@1.}
    \resizebox{\linewidth}{!}{%
    \begin{tabular}{c|c|ccc}
        \toprule
        \rowcolor{gray94}
        \textbf{Model} & \textbf{Trained on} & \textbf{NYC} & \textbf{TKY} & \textbf{CA} \\
        \midrule
        \multirow{3}{*}{LLM4POI}
            & NYC & 0.3372 & 0.2594 & 0.1885 \\
            & TKY & 0.3463 & 0.3035 & 0.1960 \\
            & CA  & 0.3344 & 0.2600 & 0.2065 \\
        \midrule
        \multirow{3}{*}{GA-LLM}
            & NYC & 0.3826 & 0.3018 & 0.2053 \\
            & TKY & 0.4059 & 0.3429 & 0.2273 \\
            & CA  & 0.3670 & 0.3065 & 0.2499 \\
        \midrule
        \multirow{3}{*}{Ours}
            & NYC & \textbf{0.4478} & \textbf{0.3253} & \textbf{0.2556} \\
            & TKY & \textbf{0.4202} & \textbf{0.3864} & \textbf{0.2650} \\
            & CA  & \textbf{0.4299} & \textbf{0.3294} & \textbf{0.3149} \\
        \bottomrule
    \end{tabular}
    }
    \label{tab:generalization}
\end{table}
To evaluate cross city generalization, we train the model on one city and infer directly on the other two unseen cities. As shown in Table~\ref{tab:generalization}, our method consistently achieves the best transfer performance. We attribute this to the shared SID token space, including geospatial prefix and semantic anchor sub-tokens that are drawn from a global, city agnostic vocabulary and have been observed during pretraining. Consequently, cross city transfer does not require extrapolating to unseen symbols, but instead amounts to recombining familiar spatial and semantic primitives in new configurations. Mobility CoT and spatial-guided RL further encourage this compositional, geography-aware behavior, enabling the model to learn transferable mobility regularities instead of memorizing city-specific trajectories.

\paragraph{Effect of History Sequence Length}
\begin{table}[t]
    \centering
    \caption{\textbf{Effect of history sequence length.}
    HR@1 with different maximum trajectory history lengths.}

    \resizebox{0.8\linewidth}{!}{%
    \begin{tabular}{c|RRR}
        \toprule
        \rowcolor{gray94}
        \textbf{History Length} & \textbf{NYC} & \textbf{TKY} & \textbf{CA} \\
        \midrule
        30  & 0.4153 & 0.3403 & 0.2835 \\
        50  & 0.4202 & 0.3626 & 0.3024 \\
        100 & 0.4416 & 0.3742 & 0.3013 \\
        300 & \bcell{\textbf{0.4478}} & \bcell{\textbf{0.3864}} & \bcell{\textbf{0.3149}} \\
        \bottomrule
    \end{tabular}
    }

    \label{tab:history-length}
\end{table}
We further analyze the effect of trajectory history length, see Table~\ref{tab:history-length}. Increasing the available history sequence length consistently improve the model's performance across all three datasets. Notably, the gains exhibit a clear marginal effect, when the trajectory history is short, extending the history leads to substantial performance improvements, while further increasing the history length beyond a moderate range yields only limited additional gains. Suggesting that most informative mobility signals are concentrated in recent and mid-range check-ins, whereas very long histories mainly provide redundant or weakly relevant information.

\subsection{Analysis}
\paragraph{Geographic Error Distribution}
Beyond accuracy, we evaluate the geographic plausibility of next POI recommendation by measuring the spatial deviation between the predicted POI and the ground-truth POI. For each test instance, we compute the Haversine distance error from the GPS coordinates of the predicted and true POIs, and report the cumulative distribution function (CDF) of distance errors in Figure~\ref{fig:dist-cdf}. A method whose CDF rises faster yields smaller spatial errors for a larger fraction of cases. As shown, our approach consistently dominates the baseline across all three cities, indicating more localized and geographically feasible recommendations. In particular, the 50th, 75th, and 90th percentile distance errors are consistently reduced across datasets, suggesting that incorporating geographic reasoning not only improves ranking metrics, but also mitigates large-distance mismatches and better aligns recommendations with real world human mobility patterns.

\paragraph{Top-K Recommendation via Two-Stage Inference.}
\begin{table*}[t]
    \centering
    \caption{\textbf{Top-K evaluation on NYC.} HR@K (Recall@K) and NDCG@K with $K\in\{5,10,20\}$.}
    \label{tab:topk-nyc}
    \resizebox{\linewidth}{!}{%
    \begin{tabular}{l|ccccc|cc|ccc>{\columncolor{blue!10}}c}
        \toprule
        \rowcolor{gray94}
        \multicolumn{1}{c|}{} &
        \multicolumn{5}{c|}{\textbf{Traditional}} &
        \multicolumn{2}{c|}{\textbf{POI Related}} &
        \multicolumn{4}{c}{\textbf{Generative}} \\
        \rowcolor{gray94}
        \multicolumn{1}{c|}{\multirow{-2}{*}{\textbf{Metric}}} &
        \textbf{SASRec} & \textbf{BERT4Rec} & \textbf{GRU4Rec} & \textbf{Caser} & \textbf{S$^3$-Rec} &
        \textbf{TPG} & \textbf{Rotan} & \textbf{TIGER} &
        \textbf{GNPR-SID} & \textbf{OneLoc} & \textbf{Ours} \\
        \midrule
        HR@5   & 0.3151 & 0.2857 & 0.1977 & 0.2883 & 0.3071 & 0.3551 & 0.4448 & 0.4965 & 0.5311 & \underline{0.6107} & \bcell{\textbf{0.6372}} \\
        HR@10  & 0.3896 & 0.3564 & 0.2460 & 0.3570 & 0.3854 & 0.4441 & 0.5223 & 0.5514 & 0.5942 & \underline{0.6563} & \bcell{\textbf{0.6897}} \\
        HR@20  & 0.4506 & 0.4130 & 0.2889 & 0.4135 & 0.4503 & 0.5121 & 0.5834 & 0.6001 & 0.6455 & \underline{0.6977} & \bcell{\textbf{0.7284}} \\
        NDCG@5 & 0.2224 & 0.2074 & 0.1442 & 0.2044 & 0.2235 & 0.2464 & 0.3471 & 0.4131 & 0.4430 & \underline{0.5355} & \bcell{\textbf{0.5398}} \\
        NDCG@10& 0.2467 & 0.2304 & 0.1599 & 0.2267 & 0.2489 & 0.2755 & 0.3723 & 0.4276 & 0.4634 & \underline{0.5504} & \bcell{\textbf{0.5570}} \\
        NDCG@20& 0.2622 & 0.2448 & 0.1708 & 0.2410 & 0.2654 & 0.2927 & 0.3878 & 0.4443 & 0.4766 & \underline{0.5608} & \bcell{\textbf{0.5668}} \\
        \bottomrule
    \end{tabular}
    }
\end{table*}

Our main setting evaluates one-to-one next POI recommendation with HR@1. To test whether Mobility CoT also supports ranking, we introduce a lightweight Top-$K$ variant that keeps \textsc{Personality Modeling} and \textsc{Intent Space Construction} to form an evidence guided candidate set, and replaces the iterative \textsc{Locality Informed Pruning} with brief, ordered rationales that justify candidates in rank order. We then apply a two-pass inference procedure: we first generate the rationale-augmented CoT trace, and then condition on this trace to perform constrained beam search and output a ranked list. Notably, the ranked outputs often include POIs that do not appear in the user’s historical sequence, indicating improved exploration beyond repeatedly visiting the same places. As shown in Table~\ref{tab:topk-nyc}, this variant consistently improves HR and NDCG, outperforming the strongest baseline\cite{kang2018self,sun2019bert4rec,hidasi2015session,tang2018personalized,zhou2020s3,luo2023timestamps,feng2024rotan,rajput2023recommender,wang2025generative,wei2025oneloc}.

\paragraph{Analysis of CoT-Free Varient}
\begin{table}[t]
    \centering
    \caption{\textbf{HR@1 of CoT-free variant.}}
    \label{tab:cot-free}
    \resizebox{0.8\linewidth}{!}{%
    \begin{tabular}{l|ccc}
        \toprule
        \rowcolor{gray94}
        \textbf{Training Setting} & \textbf{NYC} & \textbf{TKY} & \textbf{CA} \\
        \midrule
        		
        Pretrain & 0.3918 & 0.3372 & 0.2657 \\
        ROS$^{\dagger}$  & \bcell{\textbf{0.4181}} & \bcell{\textbf{0.3527}} & \bcell{\textbf{0.2874}} \\
        \bottomrule
    \end{tabular}
    }
\end{table}

We further compare a CoT-free variant of our ROS method with other baselines, see Table~\ref{tab:cot-free}. Here we employ guided decoding to suppress the explicit think stage and let the model output only the final POI recommendation. Even without generating CoT traces at inference time, ROS$^{\dagger}$ still outperforms all baselines on all three datasets. This gap shows that the proposed spatial semantic representation, structured Mobility CoT training, and spatial-guided RL not only improve performance when reasoning is explicitly generated, but also act as an effective inductive bias that strengthens the underlying model even in a pure recommendation setting.

\section{Related Work}
\label{app:relatedwork}

\paragraph{Next POI Recommendation.}
Early approaches model sequential transitions or personalized ranking, including LSTM~\cite{hochreiter1997long} and PRME~\cite{feng2015personalized}. Later methods enhance trajectory modeling with spatiotemporal attention or graph structure, such as STAN~\cite{luo2021stan}, GetNext~\cite{yang2022getnext}, and STHGCN~\cite{yan2023spatio}. More recently, LLM-based POI recommenders describe places in natural language and generate next POI outputs, represented by LLM4POI~\cite{li2024large}, while SID further improve generalization beyond numeric IDs, as explored in GNPR-SID~\cite{wang2025generative} and CoAST~\cite{zhai2025cognitive}.

\paragraph{Generative Recommendation.}
Generative recommendation replaces retrieval-and-rank with autoregressive generation conditioned on user context~\cite{rajput2023recommender,senel2024generative,gao2024generative,qu2025tokenrec}. Recent work highlights the importance of semantic item identifiers for open-vocabulary generation~\cite{deng2025onerec,chen2024hllm,han2025mtgr,wang2025generative}. For generative next POI recommendation, geographic cues are commonly injected via coordinates and addresses in prompts or via auxiliary continuous location features~\cite{feng2024move,liu2025geography,wei2025oneloc}.

\paragraph{Chain-of-Thought.}
CoT reasoning improves LLM decision making~\cite{wei2022chain}, and structured CoT stabilizes reasoning with staged templates~\cite{wang2023plan}. Prior CoT for recommendation often uses free-form preference summaries or rationales~\cite{yue2025cot4rec,tsai2024leveraging}, which can be informative but may not enforce feasibility. In contrast, our Mobility CoT explicitly uses locality informed pruning as a hard reasoning criterion for next POI selection.

\section{Conclusion}
We present ROS, a generative next POI recommendation framework that treats geography as an explicit decision variable. ROS introduces a Hierarchical Spatial Semantic ID and a three-stage Mobility CoT paradigm to construct and verify candidates under spatial and semantic constraints, then refines the model via spatial-guided RL. Extensive experiments show that ROS consistently improves hit rate and produces more geographically plausible predictions, while maintaining strong cross-city transfer. In future work, we will explore integrating abundant transportation priors for more realistic mobility reasoning.

\section*{Limitations}

While ROS demonstrates promising advancements in next POI recommendation, it also exhibits limitations in its geo-temporal feasibility modeling. By relying on a simplifying monotonic prior that long distance transitions are unlikely within short time intervals, ROS may underweight plausible rapid long range moves enabled by transportation networks such as metro rail or highways, particularly when both origin and destination are near transit hubs. Future work could incorporate network aware impedance signals such as estimated travel time or route based distance and adopt soft feasibility constraints to better preserve probability mass for such rare but realistic transitions.

\bibliography{custom}

\appendix
\section{Appendix}
\subsection{Dataset Statistics}
\label{sec:dataset-stats}

\begin{table*}
    \centering
    \caption{\textbf{Dataset statistics after preprocessing.}
    Two consecutive check-ins are assigned to the same trajectory if their time gap is within 24 hour.}
    \label{tab:dataset-stats}
    \resizebox{0.8\linewidth}{!}{%
    \begin{tabular}{lccccccc}
        \toprule
        \textbf{Data} & \textbf{Users} & \textbf{POIs} & \textbf{All Trajs} & \textbf{Valid Trajs} & \textbf{Test Trajs} & \textbf{Category} & \textbf{Check-ins} \\
        \midrule
        NYC & 1048 & 4981 & 14130& 1486  & 1447 & 318 & 103941 \\
        TKY & 2282 & 7833  & 65499 & 7174& 7079 & 291 & 405000 \\
        CA  & 3957 & 9690 & 45123& 3744  & 2864 & 296 & 238369 \\
        \bottomrule
    \end{tabular}%
    }
\end{table*}

Table~\ref{tab:dataset-stats} summarizes dataset statistics after preprocessing. We also report the numbers of trajectories in the train/test splits used throughout our experiments. Overall, the three datasets cover diverse scales in both user activity and POI catalog size, providing a comprehensive testbed for evaluating next POI recommendation methods.

\subsection{Baselines}
\label{app:baselines}

We compare against representative baselines spanning (i) \emph{traditional} metric-embedding methods,
(ii) \emph{neural} sequence/graph models with explicit spatio-temporal inductive biases, and
(iii) \emph{LLM-based} generative recommenders that leverage textualized mobility context.

\noindent\textbf{PRME}~\cite{feng2015personalized}:
Learns a metric-embedding space where next-step transition probability increases with proximity between user and POI embeddings, encouraging geographically nearby POIs as plausible next visits.

\noindent\textbf{GETNext}~\cite{yang2022getnext}:
Constructs a global flow map from collective mobility to provide collaborative priors, and integrates it into a graph-enhanced Transformer
for modeling personalized next POI transitions.

\noindent\textbf{TPG}~\cite{luo2023timestamps}:
Treats time as a prompt-like conditioning signal and uses shifted geographic windows to capture time-specific locality,
improving temporal sensitivity in location transitions.

\noindent\textbf{MTNet}~\cite{hu2018mtnet}:
Combines memory networks with transfer learning to incorporate auxiliary textual/semantic signals,
mitigating data sparsity and improving generalization to infrequent POIs.

\noindent\textbf{STHGCN}~\cite{yan2023spatio}:
Employs hypergraph convolution to model high-order spatio-temporal relations among users, POIs, and contexts,
capturing group-level mobility patterns beyond pairwise transitions.

\noindent\textbf{ROTAN}~\cite{feng2024rotan}:
Introduces rotation-based temporal encoding to represent periodic and asymmetric temporal effects,
enhancing the modeling of time-evolving mobility dynamics.

\noindent\textbf{SpaceTime-GR}~\cite{lin2025spacetimegrspacetimeawaregenerativemodel}:
A generative POI recommender that couples spatio-temporal encoders with hierarchical geographic indexing,
aiming to produce context-consistent next POIs under structured location representations.

\noindent\textbf{LLM4POI}~\cite{li2024large}:
Reformulates trajectories as natural language sequences, enabling LLMs to exploit rich contextual cues
(e.g., time, category, and address text) for next POI generation.

\noindent\textbf{GNPR-SID}~\cite{wang2025generative}:
Uses semantic discrete POI identifiers (IDs) so that functionally similar places share token-level structure,
allowing generative models to transfer knowledge across related POIs.

\noindent\textbf{GA-LLM}~\cite{liu2025geography}:
Injects GPS coordinates and transition structure into the LLM input/representation,
strengthening spatial generalization and reducing over-reliance on pure textual priors.

\noindent\textbf{CoAST}~\cite{zhai2025cognitive}:
Performs enriched pretraining followed by instruction tuning and reinforcement learning,
aligning LLM outputs with real world constraints and improving robustness for location recommendation.

\subsection{Implementation Details}
\label{app:detail}

\paragraph{SID construction and pretraining.}
We initialize the student model from \texttt{Qwen3-4B} and apply full-parameter fine-tuning throughout. 
Each POI is mapped to a hierarchical spatial-semantic identifier $\mathrm{SID}(p)=[g(p);s(p);u(p)]$.
For the geospatial prefix $g(p)$, we convert each GPS coordinate to an S2 Cell ID at the maximum S2 level, compute the dataset-wide longest common prefix over all POI cell IDs, remove this shared prefix, and keep the next $2B$ hex digits with $B{=}2$, grouped into two byte-level locality tokens $(g_1,g_2)$ and implemented as special tokens.
For the semantic anchor $s(p)$, we encode the POI category name with \texttt{Qwen-0.6B} and take the last hidden state as the POI embedding, then quantize it via a 2-level residual quantizer with codebook size 28 per level, yielding two discrete semantic tokens $(s_1,s_2)$.
The differentiating suffix $u(p)$ is a single special token assigned sequentially within each $(g,s)$ pair; empirically at most 8 suffix types are observed across all datasets.
We pretrain the model with (i) Bidirectional text SID alignment (predicting SID from category/address text and generating text from SID), and (ii) Auto-regressive next POI sequence pretraining under the SID representation.

\paragraph{Mobility CoT supervised fine-tuning.}
We use \textit{Qwen3-235B} as teacher LLM to generate three-stage Mobility CoT traces and fine-tune the student to reproduce the full trace and output the final POI. 
During training, we truncate the trajectory history to at most 50 check-ins, while at evaluation we provide up to 300 check-ins.
The number of candidates enumerated in \textsc{Intent Space Construction} is not fixed and is decided by the model itself.
We compute transition distances between adjacent check-ins (km) and bucket them into five bins: Adjacent ($\le0.2$), Nearby ($\le1.2$), Short hop ($\le3.0$), Far ($\le10.0$), and Long ($>10.0$).
For SFT, we train for 3 epochs with global batch size 8, learning rate $1\times10^{-5}$, and 100 warmup steps.

\paragraph{Spatial-Guided RL.}
We further optimize the model with GRPO using a unified reward $r=r_{\text{fmt}}+\alpha r_{\text{acc}}+\beta r_{\text{dist}}$, where $\alpha{=}2$ and $\beta{=}1$.
For the distance reward, let $d$ be the Haversine distance (km) between predicted and ground-truth POIs, and $\tilde d=\log(1+d)$.
We use two distance thresholds $d_{\text{near}}{=}0.1$ km and $d_{\text{far}}{=}3.0$ km, and define a clipped linear reward
$r_{\text{dist}}(d)=\mathrm{clip}(1+\kappa(\tilde d-\tilde d_{\text{near}}),0,1)$ with $\kappa=(0-1)/(\tilde d_{\text{far}}-\tilde d_{\text{near}})$, so predictions within $\sim$100m receive full credit and errors beyond 3km saturate to zero.
For the hierarchical SID reward, we assign $+0.3$ if $g_1$ matches and $+0.5$ if both $(g_1,g_2)$ match, and $+0.25$ if $s_1$ matches and $+0.4$ if both $(s_1,s_2)$ match; we additionally add an exactness bonus $\lambda_u{=}0.1$ when the full SID matches exactly, implemented as
$0.3\,\mathbb{I}[g_1]+\!0.2\,\mathbb{I}[g_1\wedge g_2]+\!0.25\,\mathbb{I}[s_1]+\!0.15\,\mathbb{I}[s_1\wedge s_2]+\!0.1\,\mathbb{I}[\mathrm{SID}=\widehat{\mathrm{SID}}]$.
The format reward $r_{\text{fmt}}$ is set to 2 if a regex validator confirms the required three-stage template.
We run GRPO with 4 rollouts per prompt for 3 epochs, learning rate $1\times10^{-6}$, and PPO-style clipping $\epsilon{=}0.2$; rollout decoding uses temperature $=1.0$ and top-$p=1.0$.
\begin{figure*}[t]
    \centering
    \includegraphics[width=\linewidth]{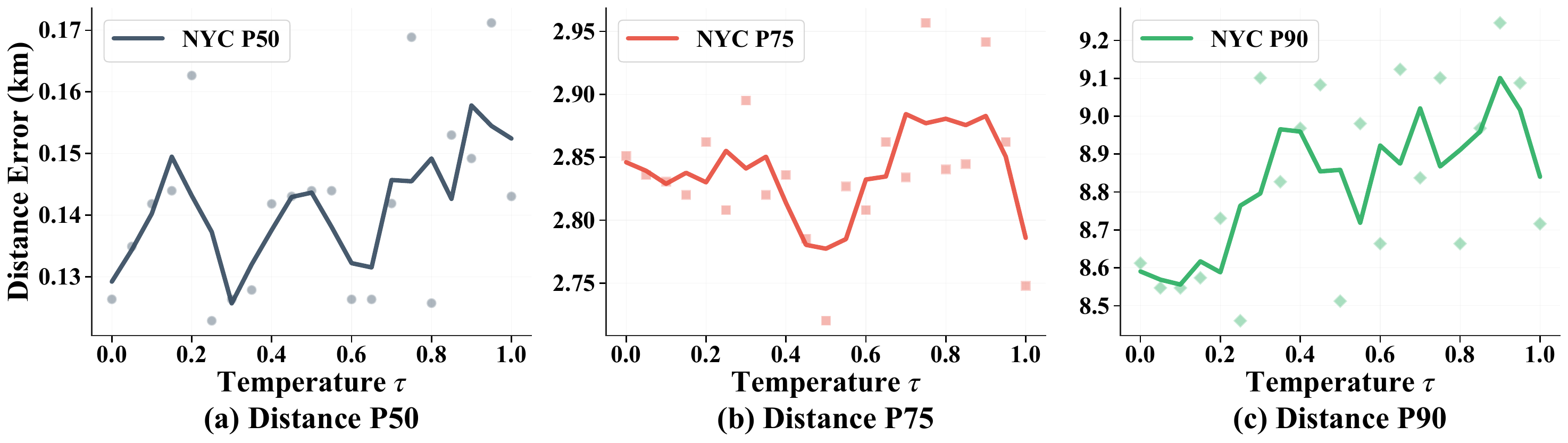}
    \caption{\textbf{NYC distance error percentiles versus sampling temperature $\tau$.}}
    \label{fig:temp-dist-nyc}
\end{figure*}

\begin{figure*}[t]
    \centering
    \includegraphics[width=\linewidth]{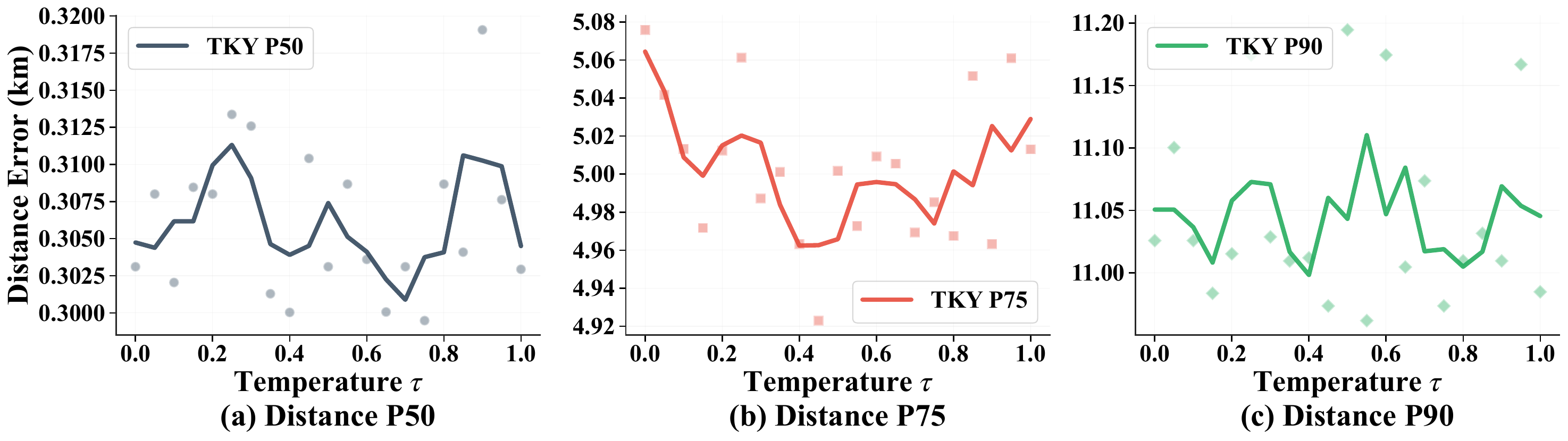}
    \caption{\textbf{TKY distance error percentiles versus sampling temperature $\tau$.}}
    \label{fig:temp-dist-tky}
\end{figure*}
\subsection{Hyperparameter Analysis}

\subsubsection{Impact of Temperature}
\begin{figure}[H]
    \centering
    \includegraphics[width=\linewidth]{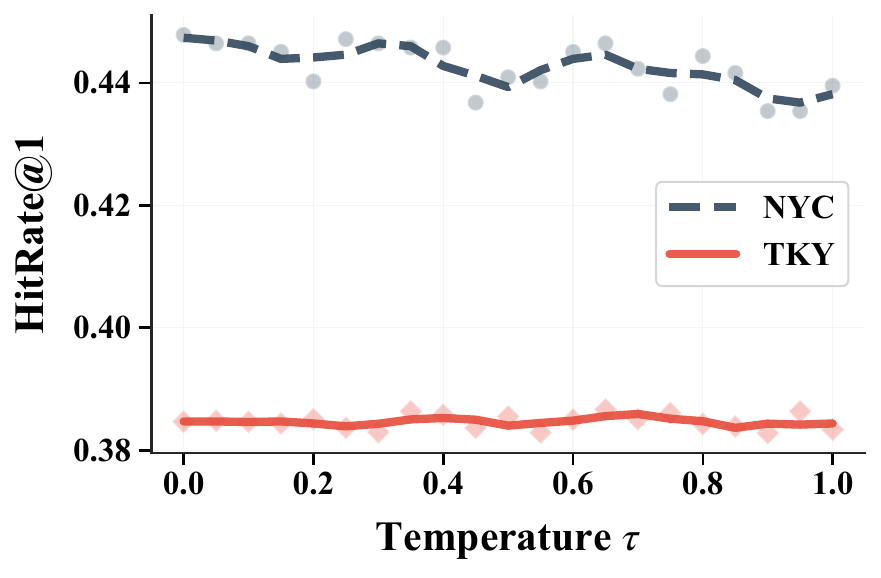}
    \caption{\textbf{HitRate@1 versus sampling temperature $\tau$ on NYC and TKY.}}
    \label{fig:temp-hitrate}
\end{figure}

As shown in Fig.~\ref{fig:temp-hitrate}, increasing the sampling temperature $\tau$ generally degrades next POI accuracy. This effect is most evident on NYC, the MA3-smoothed trend reveals a clear downward drift as $\tau$ increases, indicating that higher decoding randomness gradually pushes predictions away from the most likely destination. TKY appears less sensitive, but still shows no consistent gains from higher temperatures. Distance-error percentiles further support this observation (Figs.~\ref{fig:temp-dist-nyc} and~\ref{fig:temp-dist-tky}). As $\tau$ grows, the tail errors tend to increase or become more volatile, suggesting that higher temperature mainly introduces occasional long-range deviations rather than improving typical cases. Based on these results, we adopt deterministic decoding with $\tau=0$ for all experiments to obtain the most stable and strongest overall performance.

\subsubsection{Impact of Mobility CoT Data Scale}

\label{app:cot-scale}

\begin{figure}[H]
    \centering
    \includegraphics[width=0.9\linewidth]{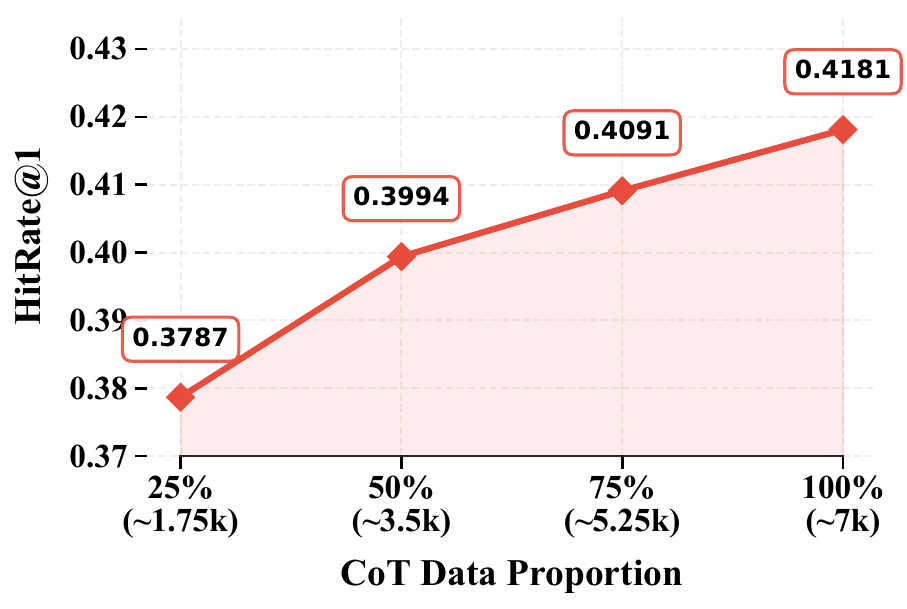}
    \caption{\textbf{HR@1 of scaling Mobility CoT supervision on NYC.}}
    \label{fig:cot-scale}
\end{figure}

To understand how much structured reasoning supervision is needed, we vary the amount of Mobility CoT traces used for NYC in the SFT stage while keeping the model, training recipe, and evaluation protocol unchanged. Specifically, we subsample the CoT training set with different ratios, where $1.0$ denotes the full set with about 7k CoT traces.

Figure~\ref{fig:cot-scale} shows a consistent monotonic improvement in HR@1 as more CoT data is used. The gain is most pronounced when increasing the ratio from 0.25 to 0.50, and gradually saturates afterwards, suggesting diminishing returns at higher supervision scales. Overall, these results indicate that Mobility CoT supervision is beneficial even at moderate scale, while using the full set provides the best accuracy in our setting.

\subsection{Case Study}
\begin{figure}[H]
    \centering
    \includegraphics[width=0.9\linewidth]{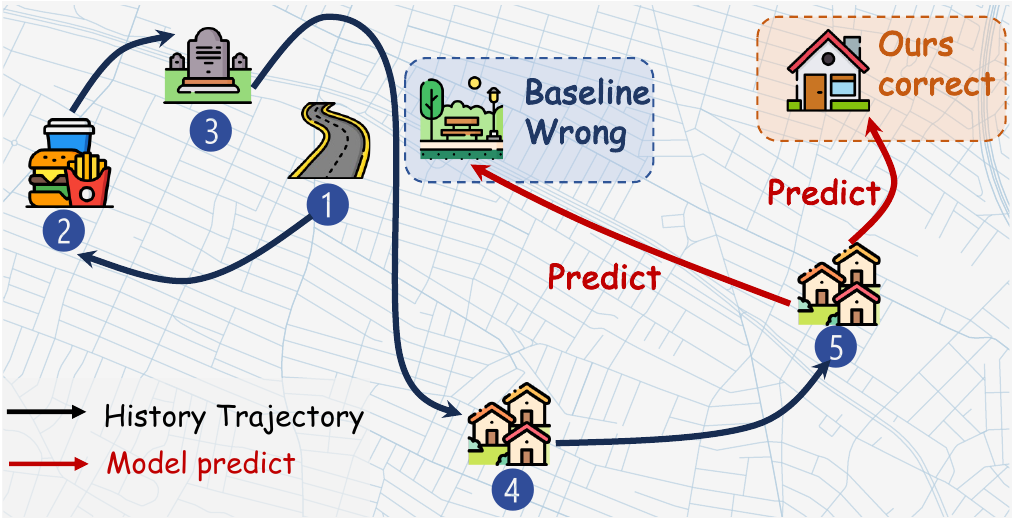}
    \caption{A representative case}
    \label{fig:case_study}
\end{figure}

Figure~\ref{fig:case_study} shows a representative case where geographic feasibility matters. The user exhibits a strong local routine, with frequent visits to a small set of nearby POIs and mostly short transitions. The task is to predict the next POI one hour after the last check-in, making long-distance moves unlikely. Mobility CoT enforces this spatiotemporal prior by restricting candidates to nearby POIs and then choosing the one most consistent with recent habits, it predicts \textit{Home} due to both proximity and the user’s typical afternoon return pattern. By contrast, the pretrain-only model recommends a POI several kilometers away, which is travel-time implausible. Mobility CoT thus acts as a feasibility filter that mitigates long-jump artifacts.
\clearpage
\onecolumn
\subsection{Prompt Format and Examples}
\newlength{\HeadBarH}
\setlength{\HeadBarH}{7mm}
\newtcblisting{HeaderListingBox}[2]{%
    breakable,
  enhanced,
  colback=white,
  colframe=#1!70!black,
  boxrule=0.8pt,
  arc=2mm,
  overlay={},
  overlay unbroken and first={%
    \begin{tcbclipinterior}
      \fill[#1] (interior.north west)
        rectangle ([yshift=-\HeadBarH]interior.north east);
    \end{tcbclipinterior}
    \node[anchor=west, font=\bfseries\color{white}]
      at ([xshift=6mm,yshift=-0.5\HeadBarH]frame.north west) {#2};
  },
  extras unbroken and first={%
    top=9mm,
  },
  listing only,
  listing options={
    basicstyle=\ttfamily\scriptsize,
    breaklines=true,
    breakatwhitespace=false,
    breakindent=0pt,
    breakautoindent=false,
    postbreak=\mbox{},
    columns=fullflexible,
    keepspaces=true,
    showstringspaces=false
  }
}
\begin{figure*}[h]
\centering
\begin{HeaderListingBox}{purple!70!black}{Input Prompt}
<|im_start|>user
Here is a record of a user's POI accesses, your task is based on the history to predict the POI that the user is likely to access at the specified time.
Given user historical data:
User Traj#1:
April 11th, 2012, Wednesday, 04:59, visit Parking at 85 Washington St <m_161><n_17><a_21><b_8><c_0>. 
April 11th, 2012, Wednesday, 14:30, visit Office at 101 Broadway <m_161><n_115><a_12><b_7><c_0>, distance is Nearby. 
April 12th, 2012, Thursday, 04:50, visit Parking at 85 Washington St <m_161><n_17><a_21><b_8><c_0>, distance is Nearby. 
April 12th, 2012, Thursday, 05:44, visit Office at 101 Broadway <m_161><n_115><a_12><b_7><c_0>, distance is Nearby. 
April 13th, 2012, Friday, 04:56, visit Parking at 85 Washington St <m_161><n_17><a_21><b_8><c_0>, distance is Nearby. 
User Traj#2:
April 19th, 2012, Thursday, 04:58, visit Parking at 85 Washington St <m_161><n_17><a_21><b_8><c_0>. 
April 19th, 2012, Thursday, 05:50, visit Office at 101 Broadway <m_161><n_115><a_12><b_7><c_0>, distance is Nearby. 
April 20th, 2012, Friday, 04:55, visit Parking at 85 Washington St <m_161><n_17><a_21><b_8><c_0>, distance is Nearby. 
April 20th, 2012, Friday, 06:07, visit Office at 101 Broadway <m_161><n_115><a_12><b_7><c_0>, distance is Nearby. 
Given user behavior sequence:
April 23rd, 2012, Monday, 04:55, visit Parking at 85 Washington St <m_161><n_17><a_21><b_8><c_0>. 
April 23rd, 2012, Monday, 05:27, visit Office at 101 Broadway <m_161><n_115><a_12><b_7><c_0>, distance is Nearby. 
April 24th, 2012, Tuesday, 04:45, visit Parking at 85 Washington St <m_161><n_17><a_21><b_8><c_0>, distance is Nearby. 
At April 24th, 2012, Tuesday, 04:58, user will visit <|im_end|>
<|im_start|>assistant
\end{HeaderListingBox}
\end{figure*}
\begin{figure*}[h]
\begin{HeaderListingBox}{purple!70!black}{Reasoning Process and Recommendation}
<think>
Step 1: On April 24th, 2012, Tuesday, the user visited Parking at 85 Washington St at 04:45. Historically, after visiting this parking location, the user consistently transitions to the Office at 101 Broadway, with visits occurring shortly thereafter.
Step 2: Possible next-POI candidates are:
1. Office at 101 Broadway, last visited on April 23rd, 2012, at 05:27.
2. Parking at 85 Washington St, last visited on April 24th, 2012, at 04:45.
Step 3: 
1. Parking at 85 Washington St: The user already visited this location on April 24th, 2012, at 04:45. Historical data shows no pattern of revisiting Parking at 85 Washington St immediately after a recent visit. 
2. Office at 101 Broadway: The user has repeatedly transitioned from Parking at 85 Washington St to the Office at 101 Broadway, especially after early morning visits, and the distance is Nearby, supporting this transition.
</think>
<m_161><n_115><a_12><b_7><c_0>
\end{HeaderListingBox}
\end{figure*}

\end{document}